\documentclass[10pt,twocolumn,letterpaper]{article}

\PassOptionsToPackage{dvipsnames}{xcolor}
\PassOptionsToPackage{table}{xcolor}

\usepackage[noindentafter]{titlesec}
\usepackage{style/cvpr}
\belowdisplayskip \abovedisplayskip

\usepackage{letltxmacro}

\LetLtxMacro{\oldsection}{\section}
\renewcommand{\section}[1]{
    \vspace{-0.11in}
    \oldsection{#1}
    \vspace{-0.10in}
}

\LetLtxMacro{\oldsubsection}{\subsection}
\renewcommand{\subsection}[1]{
    \vspace{-0.09in}
    \oldsubsection{#1}
    \vspace{-0.08in}
}

\LetLtxMacro{\oldsubsubsection}{\subsubsection}
\renewcommand{\subsubsection}[1]{
    \vspace{-0.06in}
    \oldsubsubsection{#1}
    \vspace{-0.05in}
}

\usepackage{times}
\usepackage{inconsolata}
\usepackage{xspace}

\usepackage{epsfig}
\usepackage{graphicx}
\usepackage{wrapfig}

\usepackage{cite}

\usepackage{algorithm}
\usepackage{algorithmic}

\usepackage{paralist}

\usepackage{array}
\usepackage{multirow}
\usepackage{colortbl}
\usepackage{rotating}
\usepackage{makecell}
\usepackage{booktabs}
\usepackage{tabularx}

\usepackage{amsmath, amsthm, amssymb}
\usepackage{textcomp}
\usepackage{stmaryrd}
\usepackage{upgreek}
\usepackage{bm}
\usepackage{cases}
\usepackage{mathtools}
\usepackage{arydshln} 
\usepackage{nicefrac}

\usepackage{url}
\usepackage[T1]{fontenc}
\usepackage[normalem]{ulem}
\usepackage{comment}
\usepackage{color}
\usepackage{xcolor}
\usepackage{microtype}
\usepackage{balance}
\usepackage{framed}
\usepackage{epigraph}
\usepackage{soul}
\usepackage{pifont}

\usepackage{adjustbox}%
\usepackage{textcomp}%
\usepackage{standalone}%
\usepackage{tikz}%
\usepackage{media9}

\usepackage[hang,flushmargin]{footmisc}

\usepackage{macros/mysymbols}

\newcommand{\xhdr}[1]{\vspace{2pt}\noindent\textbf{#1}}

\newcommand{\locobot}{LoCoBot\xspace}%
\newcolumntype{Y}{>{\centering\arraybackslash}X}
\newcolumntype{P}[1]{\begin{center}>{\arraybackslash}p{#1}\end{center}}

\newcounter{countitems}
\newcounter{nextitemizecount}
\newcommand{\setupcountitems}{%
  \stepcounter{nextitemizecount}%
  \setcounter{countitems}{0}%
  \preto\item{\stepcounter{countitems}}%
}
\makeatletter
\newcommand{\computecountitems}{%
  \edef\@currentlabel{\number\c@countitems}%
  \label{countitems@\number\numexpr\value{nextitemizecount}-1\relax}%
}
\newcommand{\nextitemizecount}{%
  \getrefnumber{countitems@\number\c@nextitemizecount}%
}
\newcommand{\previtemizecount}{%
  \getrefnumber{countitems@\number\numexpr\value{nextitemizecount}-1\relax}%
}
\makeatother    
\newenvironment{AutoMultiColItemize}{%
\ifnumcomp{\nextitemizecount}{>}{3}{\begin{multicols}{2}}{}%
\setupcountitems\begin{itemize}}%
{\end{itemize}%
\unskip\computecountitems\ifnumcomp{\previtemizecount}{>}{3}{\end{multicols}}{}}

\DeclareMathSymbol{@}{\mathord}{letters}{"3B}

\definecolor{citecolor}{HTML}{2980b9}
\definecolor{linkcolor}{HTML}{c0392b}
\usepackage[pagebackref=true,breaklinks=true,colorlinks,bookmarks=false,citecolor=citecolor,linkcolor=linkcolor]{hyperref}

\usepackage{cleveref}
\usepackage[font=small,labelfont=bf]{subcaption}
\usepackage{footmisc}

\usepackage{etoolbox,refcount}
\usepackage{multicol}

\begin{document}
\cvprfinalcopy 

\def\cvprPaperID{****} 
\def\httilde{\mbox{\tt\raisebox{-.5ex}{\symbol{126}}}}


\newcommand{\ourtitle}{
ObjectNav Revisited: On Evaluation of Embodied Agents Navigating to Objects
\xspace}


\title{\ourtitle}
\author{
\textbf{
Dhruv Batra$^{2,3}$, 
Aaron Gokaslan$^{2}$, 
Ani Kembhavi$^{1}$, 
Oleksandr Maksymets$^{2}$,}\\
\textbf{
Roozbeh Mottaghi$^{1}$, 
Manolis Savva$^{5,2}$, 
Alexander Toshev$^{4}$, 
Erik Wijmans$^{3,2}$} 
\\[12pt]
$^{1}$Allen Institute for Artificial Intelligence, 
$^{2}$Facebook AI Research, \\
$^{3}$Georgia Institute of Technology,
$^{4}$Robotics at Google,
$^{5}$Simon Fraser University
}


\maketitle


\begin{abstract}

We revisit the problem of Object-Goal Navigation (ObjectNav). 
In its simplest form, ObjectNav is defined as the task of navigating to an object, specified by its label, in an unexplored environment. In particular, the agent is initialized at a random location and pose in an environment and asked to 
find an instance of an object category, \eg \myquote{find a chair}, by navigating to it.

As the community begins to show increased interest in semantic goal specification for navigation tasks, 
a number of different often-inconsistent interpretations of this task are emerging.
This document summarizes the consensus recommendations of this working group on ObjectNav.
In particular, we make recommendations on subtle but important details 
of evaluation criteria (for measuring success when navigating towards a target object), 
the agent's embodiment parameters, 
and the characteristics of the environments within which the task is carried out.
Finally, we provide a detailed description of the instantiation 
of these recommendations in challenges organized at the Embodied AI workshop 
at CVPR 2020~\cite{embodiedaiworkshop}.

\end{abstract} 
\vspace{-10pt}
\section{Introduction}

\begin{figure*}[tp]
    \includegraphics[width=1\textwidth]{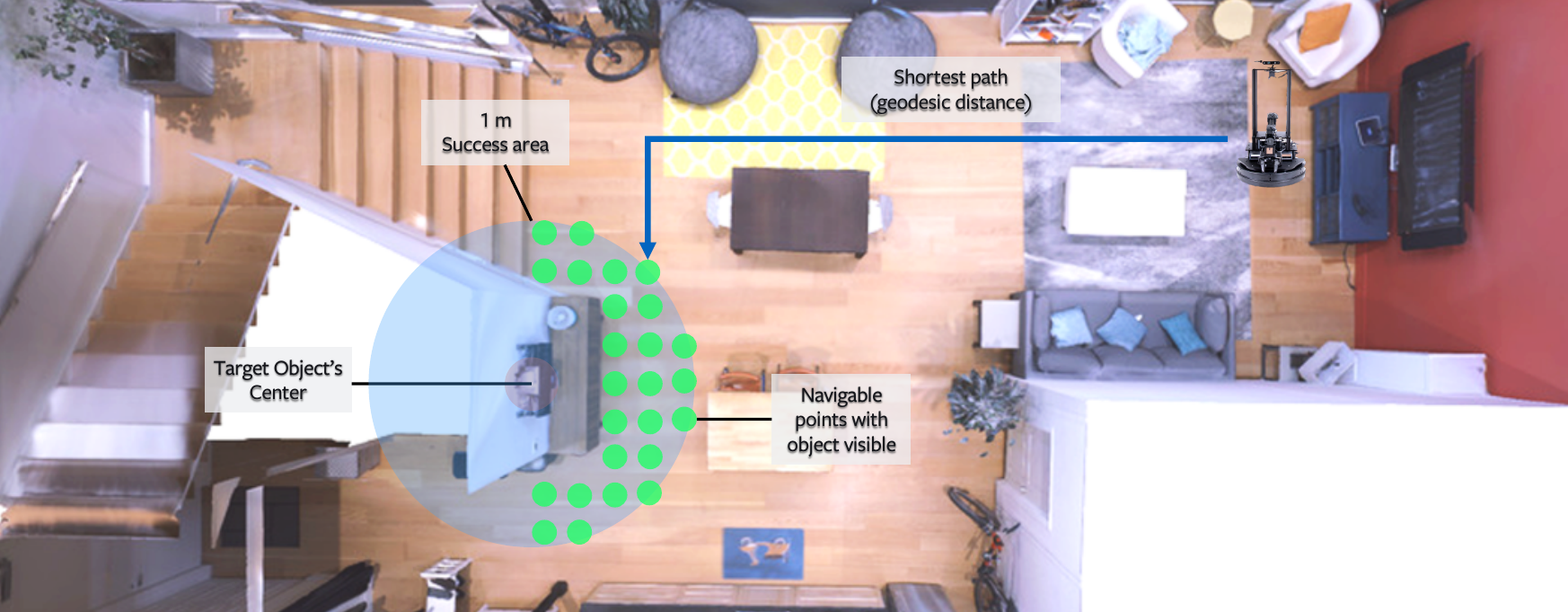}
    \caption{ObjectNav Evaluation: The goal object, in this case a monitor, is highlighted in pink. We consider all points within $1m$ to the object surface (highlighted as a blue circle) as potential success goal poses. From this area, we consider only navigable locations 
    from where the object can be seen with proper orientation of the camera (highlighted in green). These points cumulatively 
    form the `success zone', locations where if the agent called `stop' the episode would be considered successful. 
    The blue arrows shows the shortest path from the agent's (random) starting location and the closest point in 
    the success zone.}
    \label{fig:eval} 
\end{figure*}

\setlength{\epigraphwidth}{0.7\columnwidth}
\epigraph{Life is a treasure hunt.}{\textit{Olivia Wyndham}}
\vspace{-10pt}

Imagine walking up to a home robot and asking \myquote{Hey -- can you go check if my laptop is on my desk? 
And if so, bring it to me.} 
In order to be successful, such a robot would need a wide range of skills -- 
visual perception, language understanding, episodic memory construction (what is where?), 
reasoning/planning, navigation, \etc. It is clear that one essential capability 
is the ability to search for and navigate to an object (\myquote{find laptop}) -- that is the focus of this document.

\xhdr{Internet to Embodied AI.} 
Stepping back, the AI research community is undergoing a paradigm shift from tasks involving static datasets of images or 
pre-recorded experiment trials to 
tasks that rely on the deployment of an active, embodied agent in the world 
(\ie a mobile robot platform) or in simulation (\ie a virtual robot). 
Consequently, we have seen a surge of activity in Embodied AI -- 
a term referring to AI agents that act within the constraints of their embodiment in either physical 
or virtual three-dimensional environments.
This umbrella term encompasses a variety of tasks such as navigation in realistically structured interiors, 
manipulation of objects, and 
execution of natural language instructions. 

At a time of such vigorous interest, the need to unambiguously define and systematically evaluate such tasks emerges as 
a fundamental requirement for research progress. 
Anderson~\etal~\cite{anderson2018evaluation} recently 
categorized embodied navigation tasks via the description of the navigation goal: 
PointGoal (a point in space), 
ObjectGoal (a semantically distinct object instance), 
and AreaGoal (a semantically distinct area).
In addition, this effort resulted in a set of recommendations, 
such as the `success weighted by inverse path length' (SPL) metric for evaluation 
and the use of a `stop' action by agents to indicate completion of an episode.

The effort of Anderson et al.~\cite{anderson2018evaluation} was highly fruitful and led to broad 
convergence 
-- as evidenced by studies \cite{habitat19iccv, habitatsim2real19arxiv} 
and challenges~\cite{habitat_challenge} organized on the task. 
However, as the community begins to show increased interest in ObjectGoal navigation, 
a number of different often-inconsistent interpretations of this task are emerging.
In particular, there are a number of seemingly low-level but highly consequential details 
that can vary -- such as the definition of what 
constitutes success when navigating towards a target object, the details of the agent's embodiment 
parameters (in terms of possible actions and observations), and the characteristics of the environments 
within which the task is carried out.

We believe the community would benefit from finer-grained characterization of the ObjectGoal task.
Therefore, we established a multi-institutional working group of researchers in this area.   
This document summarizes our consensus recommendations. 
Specifically, we provide a set of recommendations on ObjectNav 
augmenting those in Anderson~\etal~\cite{anderson2018evaluation} and a detailed description of the instantiation 
of these recommendations in challenges organized at the Embodied AI workshop 
at CVPR 2020~\cite{embodiedaiworkshop}.

\section{ObjectNav Task Definition}
\label{sec:objectnav-task}

In its simplest form, ObjectNav is defined as the task of navigating to an object (specified by its category label) 
in an unexplored environment. In particular, the agent is initialized at a random pose in an environment and asked to 
find an instance of an object category, \eg ~\myquote{chair}, by navigating to it.

This task is ostensibly simple to define.
However, this high-level description is significantly under-specified 
and different instantiations can have different experimental outcomes.
In particular, one ought to precisely define: 
\begin{description}

 \item[Object Finding \& Evaluation:] Why are we trying to find objects and what constitutes a successful navigation?

\item[Embodiment:] What is the agent specification, in terms of actions and observations?

\item[Environments:] What environments (object categories, scenes) should be used?

\end{description}
In the following, we provide general guidelines for the above details, 
with concrete instantiations in Habitat and RoboTHOR 2020 challenges in 
\secref{sec:habitat20} and \secref{sec:robothor20} respectively. 

\subsection{Object Finding and Evaluation}

It is important to start with the question -- why is the agent trying to find objects? 
Our consensus stand is that finding (or navigating to) objects is the first step to interacting with objects. 
It is a fundamental perception and navigation milestone that forms a core building block of 
any intelligent embodied agent interacting with the world 
(opening containers, picking up objects, opening doors along the way, clearing obstacles 
out of a navigation path, responding to questions \cite{embodiedqa, iqa2018}, \etc). 
In order to enable self-contained evaluation disentangled from any specific 
downstream task, we propose evaluation protocols that measure the agent's ability to 
\emph{navigate to the object} and not object detection or framing of the object in the agent's view at the end of the episode.

At a high-level, we recommend a similar breakdown of evaluation as advocated in prior work 
\cite{anderson2018evaluation, habitat19iccv, habitat_challenge} -- 
measuring \emph{success} (did the agent navigate to the goal object?) 
and \emph{efficiency} (how efficient was the path it took compared to an optimal path?). 

These notions are reflected in our ObjectNav success criterion recommendation (see Fig.~\ref{fig:eval}). An agent has successfully navigated to the goal object iff all of the following conditions are satisfied:
\begin{itemize}
    \item \textbf{Intentionality}: The agent emits a STOP action at a location, indicating that it `believes' it has reached the object.
    \item \textbf{Validity}: The agent is at a location that is navigable given the agent's dimensions and the environmental constraints (true by construction during actual navigation but not necessarily true for a randomly sampled point on the mesh).
    \item \textbf{Proximity}: The agent's location, represented as a point (\eg center of mass), is within 
    a Euclidean distance shell around the object surface. 
    The exact size of this shell will vary depending 
    on the size of the robot and its reach; a reasonable starting point may be $r=1$m, which may be tightened over the years 
    as the community makes progress on the task.      
    The object surface may be approximated by an axis aligned or oriented 3D bounding box for computation efficacy.
    \item \textbf{Visibility}: The object is within the field of view of the agent. The exact specification of `within' 
    will depend on the agent's sensor specification (field of view, pose), actions space 
    (whether direct changes to camera roll/pitch/yaw are allowed), and the kinds of objects being considered 
    (small table-top objects or larger furniture objects). If visibility criteria is enforced, we recommend starting with 
    a fairly lenient criteria of at least 1 pixel corresponding to the object within frame, which may be tightened over the 
    years. 
    
    An alternative is to not require the agent to be actually viewing the object at the stopping location and 
    operate under \emph{oracle-visibility} -- \ie, assume access to an oracle that is able to optimally frame the object 
    in the agent's camera view without changing the agent's location -- \ie by in-place turning the agent, 
    making changes to roll/pitch/yaw of the camera, \etc. In this setting, the goal of the agent is to reach 
    locations from where the the goal object \emph{can} be viewed, but not necessarily visible from the pose the agent stops in. 
    Oracle-visibility is a proxy for 
    `the agent is close enough to interact with the object and 
    the navigation module may now hand control over to an 
    interaction module’. 
    This criterion is more suitable for scenarios that the focus is on path finding instead of both path finding and object framing. 
\end{itemize}

The blue arrow in Fig.~\ref{fig:eval} shows the shortest path from the agent's (random) starting location and the closest point in 
the success zone. ObjectNav-SPL may be defined analogous to PointNav-SPL. The key difference is that the 
shortest path is computed to the object instance closest to the agent start location. Thus, if an agent spawns 
close to \myquote{chair1} but stops at a distant \myquote{chair2}, it will succeed 
(because it found a \myquote{chair}), but at a low SPL (because the agent path is much longer compared to the oracle path).
    
Concretely, for the $i^\textrm{th}$ episode, $S_i$ denotes navigation success ($S_i=1$ if and only if the agent has succeeded as defined above), $l_i$ denotes the length of the shortest path to the closest instance of the object, and $p_i$ denotes the length of the actual path taken by the agent. The metric then reads:
\begin{align}
\textrm{SPL}_i = S_i \cdot \frac{l_i}{\max(p_i,l_i)}
\end{align}
The final evaluation metric is the average over all episodes: $\textrm{SPL} = (1/N)\sum_i\textrm{SPL}_i$.

\subsubsection{Known Issues with SPL}

\begin{table*}[t]
 \setlength{\tabcolsep}{3pt}
  \begin{center}
  \resizebox{0.9\textwidth}{!}{
    \begin{tabular}{l c c c c c c c}
      \toprule
    & \multirow{3}{*}{\parbox{1.6cm} {\centering \# of scenes}} & \multirow{3}{*}{\parbox{1.5cm} {\centering \# of target categories}} &
    \multirow{3}{*}{\parbox{2.2cm} {\centering \# of instances per category}} &
    \multirow{3}{*}{\parbox{2.0cm} {\centering \# of instances per scene per category}} & \multirow{3}{*}{\parbox{2.0cm} {\centering Sim/Real}} & \multirow{3}{*}{\parbox{2.5cm} {\centering min/max geodesic distance (m)}} & \multirow{3}{*}{\parbox{2.0cm} {\centering \# of episodes}} \\ \\ \\
      \midrule
      Habitat~\cite{habitat19iccv} & 90 & 21 & $\sim$397 & ~4 & Sim & 1/30 & 2~803K\\
      RoboTHOR~\cite{robothor} & 89 & 12 & $\sim$20 & 1 & Both & 0.71/16.8 & 34K \\
      \bottomrule
    \end{tabular}
    }
    \caption{\textbf{Environments.} 
    We provide statistics for different environments that are used for ObjectNav challenges.} 
    \label{tab:comparison}
  \end{center}
\end{table*}

We note that there are a few issues in the SPL metric that appear to be understood by domain experts but 
may not be widely known. We acknowledge and describe them below for the sake of broad dissemination: 

\begin{itemize}
\item \textbf{All failures are not equal.} 
A navigation episode is deemed unsuccessful if the agent stops outside a pre-determined radius $r$ around the goal. 
However, this results in no distinction between gross failures and minor mistakes, and gives no indication of partial progress. 
An agent navigating $30\text{m}$ to reach the goal and stopping $(r + \eps)$-distance from it 
vs one that refuses to move from the start position are equally unsuccessful, both achieving an SPL of $0$. 

\item \textbf{High Variance.} 
The binary nature of success introduces high variance in the average SPL computation, 
$\textrm{SPL} = (1/N)\sum_i\textrm{SPL}_i$. As a thought experiment, consider $N$ 
nearly identical episodes -- \ie, $\frac{l_i}{\max(p_i,l_i)}$ are nearly identical $\forall i$. 
However, due to (say) actuation noise, the agent's stopping behavior is stochastic 
-- sometimes the agent stops just outside the success criteria ($r + \eps$) and sometimes just 
inside ($r - \eps$); \ie,  
$S_i$ can be modeled as a Bernoulli random variable with probability $p$. 
In this scenario, the paths traversed by the agent are nearly identical so intuitively this should result in $Var(SPL) \approx 0$. 
However, in fact, SPL has a high variance of $p \cdot (1-p) \cdot \big[(1/N)\sum_i\textrm{SPL}_i\big]^2$. 
While this is an example exaggerated to illustrate a point, we do indeed observe in practice high variance 
in SPL estimates, particularly in settings with noisy actuations. This results in the need to evaluate 
on an unnecessarily large set of episodes to gain confidence in the SPL sample estimates.

\item \textbf{No penalty for turning.} 
$\textrm{SPL}_i$ only accounts for the distance traveled by the agent along the path it takes. On the one hand, this is a good 
idea because the length of this path is not tied to the agent's action space and can be compared across different robots. However, 
one potentially unintended side-effect is that $p_i$ is unaffected by an agent rotating/turning in-place. 
This means that SPL does not penalize the agent turning in place. Thus, an agent could always 
construct a panorama by turning $360^\circ$ before moving. An alternative would be to measure the number of actions, or energy used by the agent~\cite{xia2020interactive}. A set of path comparison metrics in the context of VLN are presented by Jain et al.~\cite{jain2019stay}.

\item \textbf{No cross-dataset comparison.} Different length paths provide different degrees of sloppiness available to the agent — short paths require a strict adherence to achieve high SPL, while long paths allow significant deviations while still achieving high SPL. See \eg Fig.~6 in \cite{ddppo}. Thus, SPL cannot be used for comparisons across different datasets or different portions of a dataset with significantly different optimal path lengths. A similar issue exists for other tasks such as object detection, when for example performances on small and large objects are compared.

\end{itemize}

In this document, we acknowledge these shortcomings of SPL but consider proposing a new navigation evaluation metric as orthogonal to characterizing ObjectNav.
Our recommendation is that whatever new metrics the community develops in the future should replace SPL for ObjectNav as well.

\subsection{Embodiment}
Embodiment
is an important part of the task insofar as we are to execute a series of realistic control actions to navigate to the object.
As such, we strive to have actions that do not trivialize the problem from a decision-making standpoint 
(\eg \myquote{teleport to kitchen}),  
while at the same time we leave low-level control challenges (\eg torque-based control of joints of an articulated robot) 
out of scope. 
The motivation behind this design decision is to bring focus on the visual and decision making aspects of the problem. 
In particular, we recommend using a discretized version of differential drive with a set of discrete actions.
As such, we come short of 
recommending modeling the full kinematics and dynamics of a robotic system. However, we do not exclude this possibility in future incarnations of the ObjectNav task.

On the observation side, we recommend modeling realistic sensors, as commonly used both in the robotics and computer vision
communities, without adding the challenge of multi-modal sensor fusion. 
For this, we recommend using color camera and depth sensors, which may be assumed to be synchronized, and localization 
sensors (\eg noisy egomotion sensors or GPS+Compass).

\subsection{Environments}

First and foremost, a target object is to be visually recognized while an agent is moving around the environment 
(as opposed to being perceived by non-visual sensors).
For a complete task definition, a set of object classes should be pre-specified (leaving open-world recognition 
out of scope, at least initially). 
Note that we recommend working with 
classes and not instances -- \ie \myquote{find a chair} not \myquote{find this specific chair} -- because 
this gives a more general understanding of recognition and navigation capabilities across 
common objects.
While we take no philosophical stand on outdoor vs indoor navigation, we 
start with establishing benchmarks on indoor navigation due to the datasets available 
and so that the set of object classes is 
more limited than the one used on general object detection, \eg COCO~\cite{lin2014microsoft}. 

To the extent possible, environments
should represent the diversity and complexity of the real world. 
We contend that it might be useful for sim2real transfer for the distribution of environments to be 
semantically representative of the real world -- 
\ie, object and spaces should be laid out in simulation as we might expect them to be in the real world. 
As such, we recommend using large sets of photo-realistic scans of real environments. 
These should have real sizes, layouts, and most importantly represent real interiors with high visual fidelity. 
Note that ObjectNav is performed in novel (previously unseen) environments. As such, it is important that the environments 
used for training and testing come from the same distribution of buildings, \eg homes.

All of these consideration overwhelmingly favor 3D scans of real spaces, rather than 
synthetically generated environments, often based on video game engines, such as \cite{beattie2016deepmind, wydmuch2018vizdoom, goseek}. Currently, there are several existing 3D environment datasets which are appropriate for ObjectNav.

\xhdr{Matterport3D}. 
Matterport3D (MP3D)~\cite{matterport} consists of scans of $90$ indoor spaces, 
with a standard split of train/val/test prescribed by Chang \etal \cite{matterport}. 
The scenes include not only private homes, but community and commercial buildings such as churches, universities, libraries, and office spaces. 
MP3D contains $40$ annotated categories, including objects (\eg doors), and floor-plan elements (\eg counter tops). 
A description of categories appropriate for ObjectNav is provided in \secref{sec:habitat20}. 

\xhdr{Gibson.} 
Another set of virtualized 3D spaces is Gibson~\cite{xia2018gibson}. This dataset currently contains $572$ scanned spaces, of which a subset has been semantically labeled with objects classes. In particular, Armeni et al.~\cite{armeni20193d} provide object segmentations and labels on $35$ homes. This dataset provides a $25/5/5$ split for training, validation, and testing. Note that many of the homes have multiple floors, which could be treated as disjoint spaces. 
We suggest using objects like `bed', `chair', `microwave', `refrigerator', `table', `toilet', `oven', `tv', 'sofa' as \mbox{ObjectNav} goals.

\xhdr{AI2-THOR.} 
AI2-THOR \cite{ai2thor} consists of a set of near-photo-realistic scenes that are designed by 3D artists. AI2-THOR currently includes two frameworks iTHOR and RoboTHOR. iTHOR is a synthetic platform mainly designed for interacting with objects and changing their state based on the physical properties of the objects and scenes. RoboTHOR consists of a set of scenes in simulation as well as the real world. The main purpose of RoboTHOR is to evaluate the generalization of models from simulation to the real world. In total, iTHOR and RoboTHOR currently have $209$ scenes and $115$ object categories which can be repositioned programatically to generate new configurations in the scenes. Both iTHOR and RoboTHOR have been used as platforms for the ObjectNav task \cite{yang19,wortsman19,robothor}.

\section{Habitat 2020 Challenge}
\label{sec:habitat20}

\begin{figure}
    \centering
    \includegraphics[width=20pc]{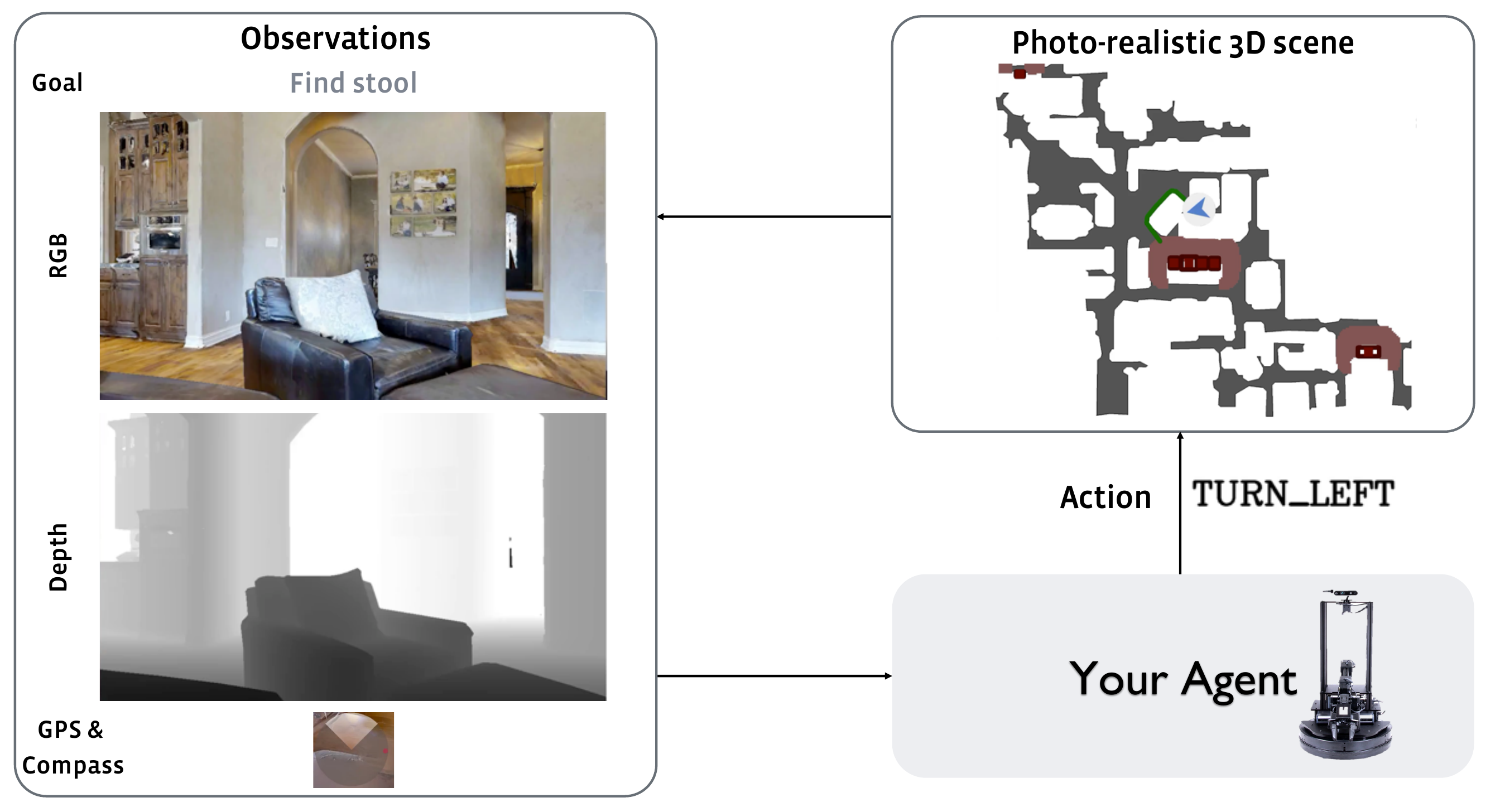}
    \caption{Habitat 2020 Challenge agent sensor and action space illustration.}
    \label{fig:habitat_objectgoal_nav_task} 
\end{figure}

\begin{figure*}[t]
    \centering
    \includegraphics[trim={79.5in 0in 0in 15in}, clip, width=\textwidth]{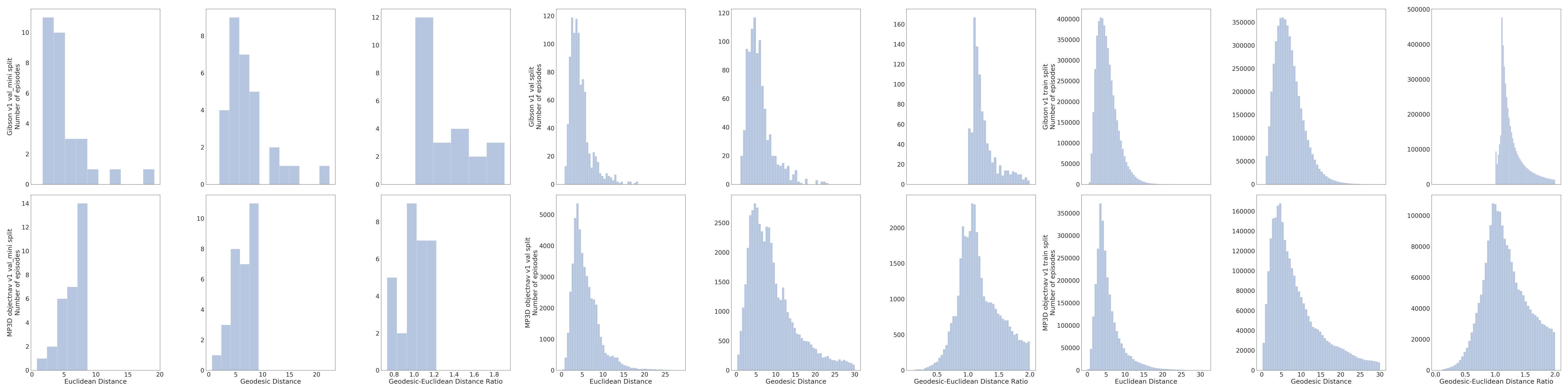}\\
    \includegraphics[trim={40in 0in 40.5in 15in}, clip, width=\textwidth]{figures/splits_stats_habitat.png}
    \caption{Statistics of ObjectNav episodes in the Habitat 2020 Challenge (top: train, bottom:val). Within each row from left: distribution over Euclidean distance between start and goal, distribution over geodesic distance along shortest path between start and goal, and distribution over the ratio of geodesic to Euclidean distance.}
    \label{fig:habitat_dataset} 
\end{figure*}

Habitat \cite{habitat19iccv} is a simulation platform for research in embodied AI. 
Habitat enables training of such embodied AI agents (virtual robots and egocentric assistants) in a 
highly photorealistic \& efficient 3D simulator, before transferring the learned skills to reality.
The Habitat challenge is an annual autonomous navigation challenge that aims 
to benchmark and accelerate progress in embodied AI.

In this section, we describe the choices made by the organizers of Habitat Challenge 2020 for the ObjectNav track. 

\noindent \textbf{Scenes.}
The Habitat 2020 challenge uses 90 scenes from the Matterport3D dataset described in \cite{matterport}. 
The choice was motivated by the desire to have photo-realistic virtualizations of real environments along with rich semantic segmentations.
These spaces have about $2$ floors and $15$ rooms, on average. 
We use the standard train/val/test splits as prescribed by Chang \etal \cite{matterport}. 
Note that as in recent works, there is no overlap between train, val, and test scenes.

\noindent \textbf{Objects.} 
As described in the previous section, MP3D contains $40$ annotated categories including objects (\eg doors), and architectural elements (\eg counter tops). 
We manually select a subset of object categories by excluding categories that are not visually well-defined (\eg doorways, windows, and architectural elements such as walls, floors, and ceilings).
In total, our subset contains $21$ categories with $8@349$ instances in $71$ environments, 
with an average of 397.5/117.59 categories/instances per environment for the training split.
We provide additional statistics across dataset splits in Figure \ref{fig:habitat_dataset}. 
The full list of object categories we use is -- `chair', `table', `picture', `cabinet', `cushion', 
`sofa', `bed', `chest\_of\_drawers', `plant', `sink', `toilet', `stool', `towel', `tv\_monitor', 
`shower', `bathtub', `counter', `fireplace', `gym\_equipment', `seating', `clothes'.

\noindent \textbf{ObjectNav Episode Dataset.} 
We create an ObjectNav episode dataset using the above categories.
An episode is defined by the unique id of the scene, 
the starting position ($x,y,z$) and orientation of the agent ($\theta$), and the goal category label (\eg `chair'). 
Additional metadata such as the Euclidean and geodesic distance along the shortest path (GDSP) 
from the start position to the goal position, 
and all valid success zone locations with viewpoints are also included. 
To determine locations that form the success zone, we follow the recommendations laid out in \secref{sec:objectnav-task} -- 
the location must be navigable (true by construction during actual navigation but not necessarily true for a 
randomly sampled point on the mesh), 
the location must be within $1$m of the oriented bounding box of the object instance or inside of the bounding box, 
and object must be visible from that location under oracle-visibility. 
To make the computation of success and distance to goal efficient, we provide such positions (in the 
success zone of each goal object) within the episode specification, and call them `valid viewpoints'.
These viewpoints were sampled with an $\frac{R_\text{agent}}{2}$ grid on the floor. 
If no locations on the floor satisfy these constraints for a particular object instance, 
we do not include that instance in our dataset. 
Finally, we restricted the maximum action length of the shortest action path to $750$ and checked the ratio of 
geodesic to Euclidean distance is at least 1.05 to rule out easy episodes.


\xhdr{Embodiment.} 
The experiments by Savva~\etal~\cite{habitat19iccv} 
and the Habitat 2019 Challenge \cite{habitat_challenge} modeled 
the agent as an idealized cylinder 
of radius ${R_\text{agent}} = 0.1\text{m}$ and height $1.5\text{m}$. 
As shown in \figref{fig:habitat_objectgoal_nav_task}, 
we configure the agent to match 
the \locobot robot as closely as possible.
Specifically, we configure the simulated agent's base radius and height to be $0.18\text{m}$ and $0.88\text{m}$ respectively.

\xhdr{Action Space.} 
The action space for the agent consists of $6$ actions: 
\texttt{move-forward $0.25\text{m}$}, 
\texttt{turn-left $30^{\circ}$}, 
\texttt{turn-right $30^{\circ}$},  
\texttt{look-up $30^{\circ}$},  
\texttt{look-down $30^{\circ}$}
and \texttt{STOP}.
Note that in contrast to \cite{habitat19iccv, habitat_challenge}, 
we increased the angles associated with \texttt{turn-left} and \texttt{turn-right} actions from $10^{\circ}$ to $30^{\circ}$ degrees due to the concerns described in \cite{habitatsim2real19arxiv}. 

\xhdr{Sensor Specification.}
The agent is equipped with an RGB-D camera and a GPS+Compass sensor; both sensors are `idealized' 
in the sense that no sensor-noise is injected beyond the reconstruction artifacts that may exist already in the underlying 
3D data.  
We attempt to match the camera specification in simulation to the Azure Kinect camera. Specifically, 
we set the camera's horizontal field of view to $79^{\circ}$, resolution to VGA (height $\times$ width = $480\times640$ pixels),  
and 
depth sensing clipped to a minimum and maximum of $0.5\text{m}$ and $6\text{m}$ respectively. 
The GPS+Compass sensor provides relative location ($x, y, z$) and heading (azimuth angle) \wrt 
the agent's start location (origin) and heading ($0^\circ$). Thus, the agent's spawn location and pose establishes a coordinate system for that episode.

\begin{figure}[t]
    \centering
    \includegraphics[width=0.95\columnwidth]{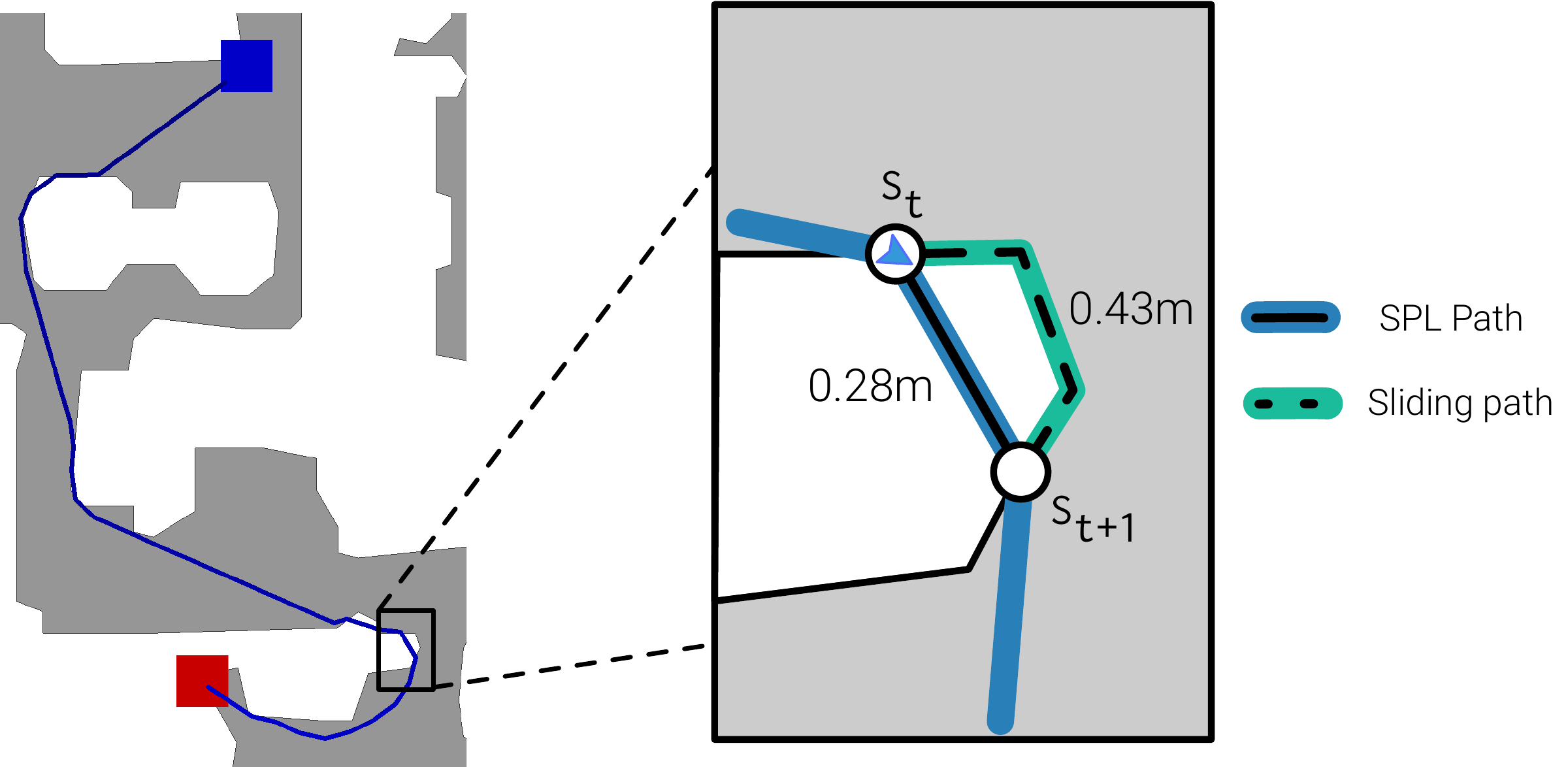}
    \caption{Sliding behavior leading to `cheating' agents. At time $t$, the agent at $s_t$ executes a \texttt{forward} action, and slides along the wall to state $s_{t+1}$. The resulting straight-line path (used to calculate SPL) goes outside the environment. Figure and caption from ~\cite{habitatsim2real19arxiv}.}
    \label{fig:corner-clipping}
\end{figure}

\xhdr{Collision Dynamics and `Sliding'.}
In prior work \cite{habitat19iccv, habitat_challenge}, 
when the agent takes an action that results in a collision, the agent \emph{slides} along the obstacle as opposed to stopping.
This behavior is prevalent in video game engines as it allows for smooth human control; it is also enabled by default in MINOS~\cite{savva2017minos}, Deepmind Lab~\cite{beattie2016deepmind}, AI2 THOR~\cite{ai2thor}, and Gibson~\cite{xia2018gibson} v1.
Recent work by Kadian et al.~\cite{habitatsim2real19arxiv} has shown that this behavior enables `cheating' by learned agents.
Specifically, as illustrated by an example in \figref{fig:corner-clipping}, the agent exploits this sliding mechanism to take an effective path that appears to travel \emph{through non-navigable regions} of the environment (like walls).
Such policies fail disastrously in the real world where the robot bump sensors  force a stop on contact with obstacles.
To rectify this issue, we configure Habitat-Sim to 
disable sliding on collisions.

\noindent \textbf{Success criteria.}
An agent navigation trajectory is considered successful if the $STOP$ action was called 
within $0.1\text{m}$ geodesic distance from a valid `viewpoint' of the goal object as defined in the Episode Dataset description (see \figref{fig:eval}). 

A complete task configuration is available at: 
\url{https://github.com/facebookresearch/habitat-challenge/blob/challenge-2020/habitat-challenge-data/challenge_objectnav2020.local.rgbd.yaml}
which we hope will help prevent inconsistencies. 

\subsection{Challenge Phases}
\noindent \textbf{Minival phase} 
The purpose of this phase is sanity checking -- to confirm that an EvalAI remote evaluation matches the results achieved by participants locally.
Each team is allowed a maximum of $30$ submissions per day for this phase.

\noindent \textbf{Test Standard phase}
The purpose of this phase is to serve as the public leaderboard establishing the state of the art.
This is what should be used to report results in papers. 
Each team is allowed a maximum of $10$ submissions per day for this phase.

\noindent \textbf{Test Challenge phase}
This phase will be used to decide challenge winners. Each team is allowed a total of $5$ submissions until the end of the challenge submission phase.
Results on this split will be made public at the Embodied AI workshop at CVPR.

\section{RoboTHOR 2020 Challenge}
\label{sec:robothor20}

\begin{figure}[tp]
    \centering
    \includegraphics[width=20pc]{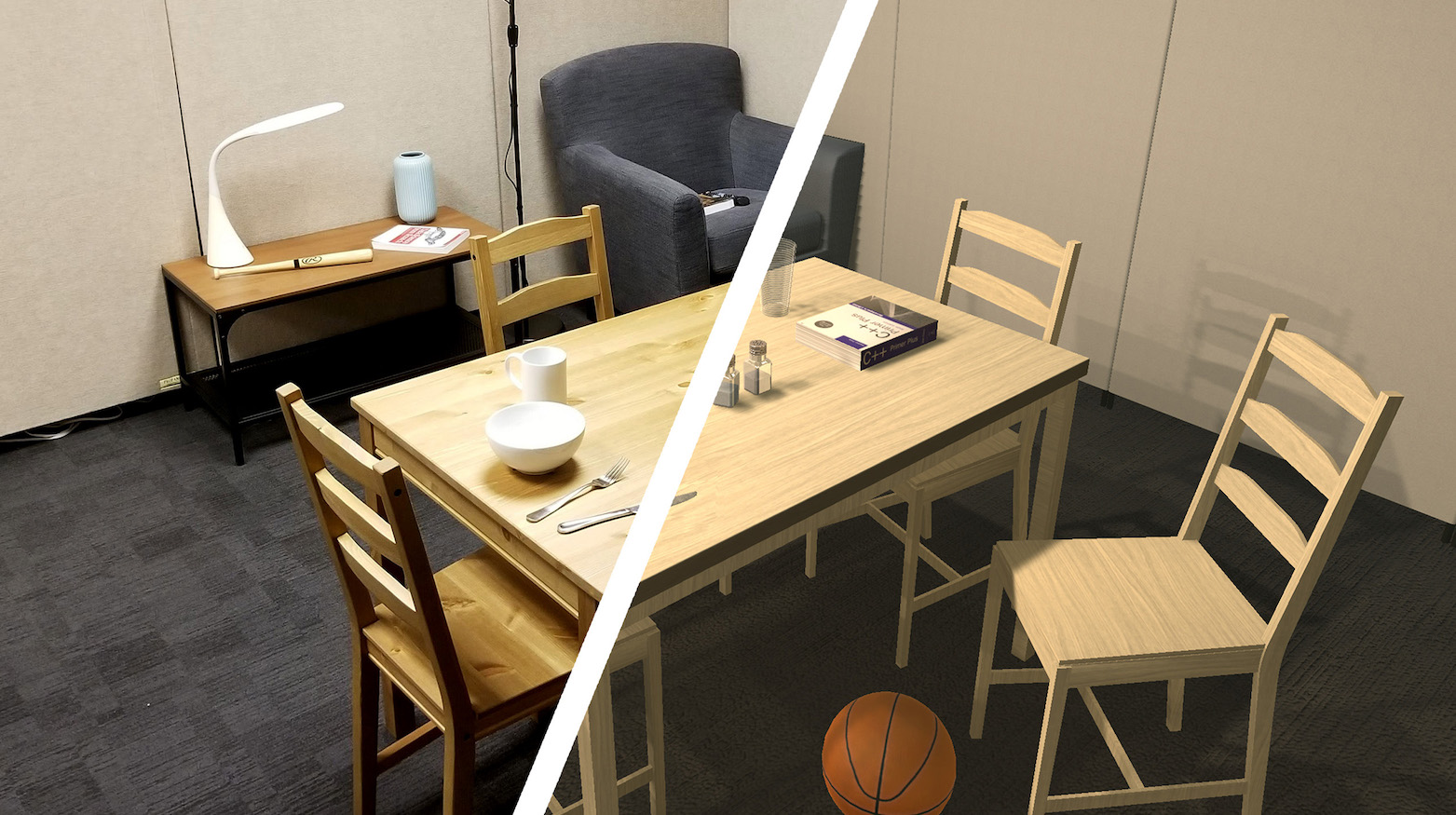}
    \caption{RoboTHOR scenes in simulation and the physical world have identical object layouts. The disparities in appearance and control dynamics make the transfer from simulation to the real world challenging.}
    \label{fig:robothor} 
\end{figure}

RoboTHOR \cite{robothor} is a platform to develop and test embodied AI agents with corresponding environments in simulation and the physical world (Fig.~\ref{fig:robothor}). The aim of the RoboTHOR 2020 challenge is to develop methods in simulation that generalize to real world environments. The complexity of environments in RoboTHOR along with disparities in appearance and control dynamics between simulation and reality pose significant and novel challenges to participants.

\begin{figure*}
    \centering
    \includegraphics[width=1\textwidth]{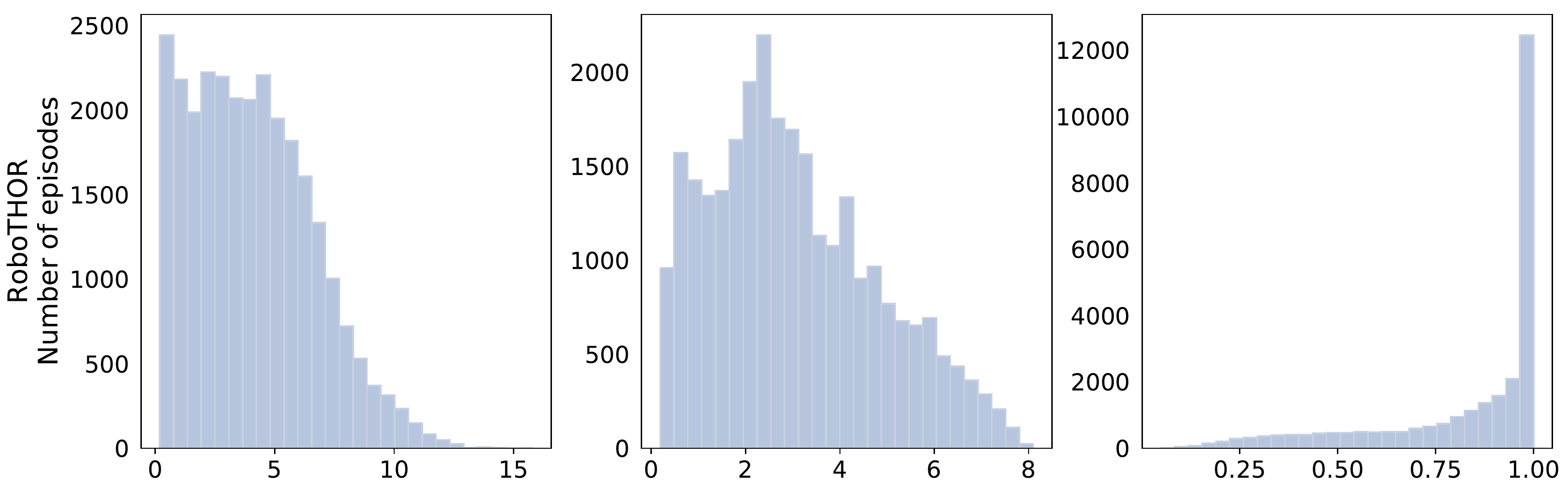}
    \caption{\textbf{Target distance distribution in RoboTHOR.} The distribution of Geodesic distances (left), Euclidean distances (middle) and their ratio (right) from the start locations to the targets.}
    \label{fig:robothor_stat} 
\end{figure*}

\noindent \textbf{Scenes.} RoboTHOR includes $89$ scenes in total: $75$ simulated scenes used for training, $4$ real/simulated scenes (referred to as test-dev) used for validation, and $10$ real scenes used for the challenge (referred to as test-challenge). While the overall dimensions of all simulation environments in RoboTHOR is consistent, there is no overlap between the interior wall layouts across these three sets. This imposes a further challenge on agents - generalizing to unseen layouts in the validation and challenge scenes.

\noindent \textbf{Objects.} Scenes in RoboTHOR are populated with objects drawn from an object asset library. There are $731$ unique object instances in RoboTHOR across $43$ object categories: $11$ classes of furniture (\eg sofa and TV stand) and $32$ classes of small objects (\eg mug and laptop). $14$ of these $43$ categories are guaranteed to be found in all scenes. We use $12$ of these categories as our test targets (we ignore very small objects such as remote control). We distribute the objects in the scenes as uniformly as possible to reduce the location bias of objects. Importantly, object instances in the train, validation and challenge sets have no overlap. This requires agents to generalize to unseen object instances when evaluating on the validation and challenge environments. 

\noindent \textbf{Agent.} We use the LoCoBot robot for this challenge with a few physical modifications. We made the agent taller ($0.88\text{m}$) to be able to observe objects on the tables and high surfaces. We have also replaced the camera by an Azure Kinect camera which has a wider field of view of $79^{\circ}$ and captures higher quality images. The agent in simulation exactly matches the modified LoCoBot. The API for controlling the robot is identical across both simulated and the real platforms. The sensors available during training are RGB and depth images as well as localization and orientation sensors. We consider two sensor settings during evaluation: (1) RGB image only and (2) RGB and depth images. The action set is identical to the one defined in Section~\ref{sec:habitat20}.

\noindent \textbf{Object visibility.} We consider a navigation episode successful if the target object is within 1$m$ and is in the field of view of the camera. In simulation this information is provided by the agent controller. In the real experiments, we localize the agent by a set of  Super-NIA-3D localization modules to measure the distance and manually check if the object is visible in the image.

\subsection{Challenge Phases}
The RoboTHOR challenge has three phases: (1) Train in simulation (2) Test-Dev in simulation and real (3) Test-challenge in real. In all phases, the robot starts at a randomly chosen location and it is provided with the target object category.

\noindent \textbf{Phase 1.} This phase involves only training using the $75$ simulated training scenes. In our setup, $60$ scenes are used for training the models, and the rest of the scenes are used for validation. We provide a dataset of starting points and targets. The dataset is distributed across easy, medium and hard cases, where the difficulty is chosen based on the geodesic distance to the target. The distribution of the distances are shown in Fig.~\ref{fig:robothor_stat}. The participants can use these points or their generated points for training.

\noindent \textbf{Phase 2.} This phase involves testing the models on $4$ scenes in the physical space and their counterparts in simulation. This enables the participants to check how well the discrepancies in simulation and the real world are handled by their model.

\noindent \textbf{Phase 3.} The models are tested on a blind test set of $10$ real scenes. These scenes do not have any overlap with the training and validation scenes in terms of wall layout and object instances. The target categories are the same as the ones used for training and validation. This phase evaluates how well the models generalize (1) from simulation to real, (2) to unseen scenes and targets. 

\section{Conclusion}
Our goal is to provide a common task framework for the task of visual navigation towards objects (ObjectNav). 
We discuss the evaluation metric, agent and environments specifications 
for this task and provide a set of recommendations for each aspect of the ObjectNav problem. 
We hope these recommendations promote consistency in future research and encourage more systematic 
benchmarking in this domain.


\vspace{1em}

\LetLtxMacro{\section}{\oldsection}
{\small
\bibliographystyle{style/ieeetr_fullname}
\bibliography{bib/strings,bib/main}
}


\end{document}